\documentclass{article}

\usepackage{ijcai16}

\usepackage{times}
\usepackage{epsfig}
\usepackage{graphicx}
\usepackage{amsmath}
\usepackage{amssymb}
\usepackage{mathsymb}
\usepackage{textcomp}
\usepackage{verbatim}
\usepackage{subfigure}
\usepackage{url}
\usepackage{graphicx}
\usepackage{multirow}
\usepackage[super]{nth}
\usepackage{balance}
\usepackage{colortbl}
\graphicspath{{./}{./Fig/}{./Figs/}{./Fig/eps/}{./Fig/pdf/}}

\usepackage{algorithm2e}

\let\savedalgorithm\algorithm
\let\savedendalgorithm\endalgorithm

\usepackage{algorithm}

\newcommand{\etal}{\textit{et al.\ }}
\newcommand{\eg}{\emph{e.g.,\ }}
\newcommand{\ie}{\emph{i.e.,\ }}

\title{Iterative Views Agreement: An Iterative Low-Rank based Structured Optimization Method to Multi-View Spectral Clustering}
\author{Yang Wang$^{\dag}$, Wenjie Zhang$^{\dag}$, Lin Wu$^{\ddag}$$^{\sharp}$, Xuemin Lin$^{\dag}$, Meng Fang$^{\S}$, Shirui Pan$^{\natural}$\\
 $^{\dag}$The University of New South Wales, Kensington, Sydney, Australia\\
 $^{\ddag}$The University of Adelaide, Australia $^{\sharp}$Australian Centre for Robotic Vision\\
 $^{\S}$The University of Melbourne, Australia $^{\sharp}$University of Technology Sydney, Australia\\
 \{wangy,zhangw,lxue\}@cse.unsw.edu.au; lin.wu@adelaide.edu.au\\
 meng.fang@unimelb.edu.au; shirui.pan@uts.edu.au
}

\begin{document}

\maketitle

\begin{abstract}

Multi-view spectral clustering, which aims at yielding an agreement or consensus data objects grouping across multi-views with their graph laplacian matrices, is a fundamental clustering problem. Among the existing methods, Low-Rank Representation (LRR) based method is quite superior in terms of its effectiveness, intuitiveness and robustness to noise corruptions. However, it aggressively tries to learn a common low-dimensional subspace for multi-view data, while inattentively ignoring the local manifold structure in each view, which is critically important to the spectral clustering; worse still, the low-rank minimization is enforced to achieve the data correlation consensus among all views, failing to flexibly preserve the local manifold structure for each view. In this paper, 1) we propose a multi-graph laplacian regularized LRR with each graph laplacian corresponding to one view to characterize its local manifold structure. 2) Instead of directly enforcing the low-rank minimization among all views for correlation consensus, we separately impose low-rank constraint on each view, coupled with a mutual structural consensus constraint, where it is able to not only well preserve the local manifold structure but also serve as a constraint for that from other views, which iteratively makes the views more agreeable.
Extensive experiments on real-world multi-view data sets demonstrate its superiority.

\end{abstract}

\section{Introduction}
Spectral clustering \cite{nips01,nips04}, which aims at  exploring the local nonlinear manifold (spectral graph) structure inherently embedded in high-dimensional data to partition data into disjoint meaningful groups, is a fundamental clustering problem. Because of its elegance, efficiency and good performance, spectral clustering has become one of the most popular clustering methods.
Recently, great attention have shifted from conventional single view/graph to multi-view spectral clustering, with the motivation of leveraging the complementary information from multi-view data sources where the same data set can be decomposed by different features \eg an image can be described by its color histogram or shape feature; one document can be represented by page link or document text. As explicitly claimed by numerous pieces of multi-view research \cite{TaoPAMIA,MVL13,WangTNN2015,YangMM14}, an individual view is unlikely to be sufficiently faithful for effective multi-view learning. Therefore, the integration of multi-view information is both valuable and necessary.

\subsection{Motivation}
Essentially, the critical issue of multi-view learning is to achieve the agreement/consensus \cite{TaoTIP14,WangTIP2015,YangPAKDD14,YangKAIS16} among all views given the complementary information from multi-views to yield a substantial superior performance in clustering over the single view paradigm. Numerous multi-view based methods are proposed for spectral clustering. \cite{CVPR12,M-V-C} performs multi-view information incorporation into the clustering process by optimizing certain objective loss function. \emph{Late fusion} strategy \cite{ecmlpkdd09} designed for multi-view spectral clustering works by first deriving the spectral clustering performance regarding each view, and then combining multiple view-induced results into an optimal one. Such strategy, however, cannot ideally achieve the multi-view agreement, as each view cannot co-regularize with each other during the clustering process.

Canonical Correlation Analysis (CCA) based methods \cite{CSC,ICML09} for multi-view spectral clustering are developed by projecting the multi-view data sources into one common  lower dimensional subspace, where the spectral clustering is subsequently conducted. One limitation of such method lies in the fact that one common lower-dimensional subspace cannot flexibly characterize the local spectral graph structures from heterogeneous views, resulting into an inferior multi-view spectral clustering. Kumar \etal \cite{NIPS11} proposed a state-of-the-art co-regularized spectral clustering for multi-view data. They attempted to regularize the eigenvectors of view-dependent graph laplacians and achieve consensus clusters across views. Similarly, a co-training \cite{COLT98,ICML10} framework is proposed for multi-view spectral clustering \cite{icml11}, where the similarity matrix from one view is projected into the subspaces spanned by the eigenvectors from other views, then spectral clustering is conducted on such projected similarity matrix.  This process is alternately performed until convergence, and the final result is formed by aggregating the clustering results from each individual view.

The above co-regularized \cite{NIPS11} and co-training \cite{icml11} based methods can effectively achieve the clustering consensus under the scenario with noise corruption free in view-dependent feature representations. However, such assumption is hard to be satisfied  in practice. To address such stand-out limitation, Low-Rank Representation (LRR) \cite{RMVSC,LRRICML2010,ICCV2011,LiuPAMI13} based approaches have been proposed for multi-view spectral clustering. The basic idea is to decompose data representation from any view into  a view-dependent noise corruption term and a common low rank based representation shared by all views, which further leads to common data affinity matrix for clustering.
The typical LRR \cite{RMVSC,LRRICML2010} model is formulated below.
\begin{equation}\label{eq:intro}
\begin{aligned}
& \min_{Z, E_i} ||Z||_{*} + \lambda\sum_{i \in V}||E_i||_1\\
& \textmd{s.t.} ~~~~~ X_i = X_iZ(or~~Z~~only) + E_i, i = 1,\ldots,V,
\end{aligned}
\end{equation}
where $V$ denotes the number of all views; $X_i \in \mathbb{R}^{d_i \times n}$ denotes the data feature representation for the $i^{th}$ view, $n$ is the number of data objects for each view; $d_i$ is the feature representation dimensions for the $i^{th}$ view.  $Z \in \mathbb{R}^{n \times n}$ represents the self-expressive linear sample correlations \cite{LRRICML2010} shared by all views with the assumption that the similar samples can be linearly reconstructed by each other.  $E_i \in \mathbb{R}^{d_i \times n}$ models the possible noise corruptions in the feature representations for the $i^{th}$ view. $||E_i||_1$ is the $\ell_1$ norm of $E_i$ representing the summation of the absolute value of all entries from $E_i$; $\lambda$ is the balance parameter.

Despite the effectiveness of LRR for multi-view spectral clustering, they still arguably face the following fundamental limitations:
\begin{itemize}
\item LRR attempts to learn a common lowest-rank representation revealing a low-dimensional subspace structure, but inattentively ignore the distinct manifold structures in each view, which turns out to be critically important to multi-view spectral clustering.

\item Low-rank constraint is imposed to enforce all views to share the consensus $Z$ in Eq.\eqref{eq:intro} \cite{RMVSC}, however, such enforced common representation may not flexibly preserve the local manifold structure from heterogeneous views, resulting into a non-ideal multi-view clustering performance.
\end{itemize}

\subsection{Our contributions}
To address those stand-out limitations, our method delivers the following novel features:

\begin{itemize}
\item To characterize the non-linear spectral graph structure from each view, inspired by \cite{PAMI16}, we propose to couple LRR with multi-graph regularization, where each graph laplacian regularization can characterize the view-dependent non-linear local data similarity.

\item To achieve the view agreement while preserving the data correlations within each view, we present an iterative view agreement process in optimizing our objective function. During each iteration, the low-rank representation yielded from each view serves as the constraint to regulate the representation learning from other views. This process iteratively boosts these representations to be more agreeable.

\item To model the above intuitions, we figure out a novel objective function and the Linearized Alternating Direction Method with Adaptive Penalty (LADMAP) \cite{LinNIPS2011} method is deployed to solve it.
\end{itemize}

\section{Iterative Low-Rank based Structured Optimization Method to Multi-view Spectral Clustering}
It is well known that the critical issue for spectral clustering lies in how to effectively model the local nonlinear manifold structure \cite{nips04}. Hence, for each view, we aim at preserving such nonlinear manifold structure of original high-dimensional data set within the space spanned by the low-rank sparse representations $Z_i$ for the $i^{th}$ view. This can be formulated as:
\begin{equation}\label{eq:spectral}
\begin{aligned}
& \frac{1}{2}\sum_{j,k}^{n}||z_j^i - z_k^i||^2W_{jk}^i \\
& = \sum_{j = 1}^{N}(z_j^i)^Tz_j^iD_{jj}^i - \sum_{j,k}^{N}(z_k^i)^Tz_j^iW_{jk}^i \\
& = \textmd{Tr}(Z_i^TD_iZ_i) - \textmd{Tr}(Z_i^TW_iZ_i) = \textmd{Tr}(Z_i^TL_iZ_i),
\end{aligned}
\end{equation}
where $z_k^i \in \mathbb{R}^{n}$ is the $k^{th}$ row of $Z_i \in \mathbb{R}^{n \times n}$ representing the linear correlation representation between $x_k$ and $x_j(j \neq k)$ in the $i^{th}$ view;  $W_{jk}^i$ is the $(j,k)^{th}$ entry of the similarity matrix $W_i$, which encodes the similarity between $x_j$ and $x_k$ from the original high dimensional space for the $i^{th}$ view; $W_i \in \mathbb{R}^{n \times n}$ is the similarity matrix for all the data objects from $X_i$; $D_i$ is a diagonal matrix with its $k^{th}$ diagonal entry to be the summation of the $k^{th}$ row of $W_i$, and $L_i = D_i - W_i$ is the graph laplacian matrix for the $i^{th}$ view; thus Eq.\eqref{eq:spectral} is always dubbed \emph{graph laplacian regularizer}. In this paper, we choose Gaussian kernel to calculate $W_{jk}^i$ as
\begin{equation}\label{eq:gaussian}
W_{jk}^i = e^{-\frac{||x_j^i - x_k^i ||^2_2}{2\sigma^2}},
\end{equation}
where $\sigma$ is the bandwidth parameter and $||\cdot||_2$ denotes the $\ell_2$ norm; Eq.\eqref{eq:gaussian} holds if $x_k^i$ is within the $s$ nearest neighbors of $x_j^i$ or vice versa, and it is 0 otherwise. $W_{jj}^i (\forall j)$ is set to 0 to avoid self-loop. Eq.\eqref{eq:spectral} explicitly requires $Z_i$ to well characterize the local manifold structure inherently embedded in original high-dimensional $X_i$ for the $i^{th}$ view, which is of importance to spectral clustering.

Based on the above, we leverage the above graph laplacian regularizer with the low-rank representation. Considering the global clustering structure captured by low-rank representation may prevent us from directly imposing graph Laplacian regularizer for local manifold structure, we propose to impose the sparsity norm $\ell_1$ on $Z_i$, denoted as $||Z_i||_1$, which can discriminatively extract the local sparse representative neighborhood of each data object.

As explicitly revealed by most of the multi-view clustering research \cite{NIPS11,icml11,M-V-C}, it is always anticipated that a data point should be assigned to the same cluster irrespective of views. In other words,
the critical issue to ensure ideal multi-view clustering performance is to achieve the clustering agreement among all views. Based on that, we aim to minimizing the difference of such low-rank and sparse representations from different views by proposing a consensus term to coordinate all views to reach clustering agreement.
\begin{equation}\label{eq:object}
\begin{aligned}
& \min_{Z_i, E_i (i \in V)} \sum_{i \in V} ( \underbrace{||Z_i||_*}_\text{Low-rank representation} + \underbrace{\lambda_{1} ||E_i||_1}_\text{noise and corruption robustness}\\
&+\underbrace{\lambda_{2}||Z_i||_1}_\text{local sparsity modeling} + \underbrace{\lambda_{3}\textmd{Tr}(Z_i^TL_iZ_i)}_\text{Graph regularization}\\ &+ \underbrace{\frac{\beta}{2}\sum_{j \in V, j \neq i}||Z_i - Z_j||^2_2)}_\text{Views-agreement}\\
&~~~~~~~~~~~\textmd{s.t.}~~~~ i = 1, \ldots, V, ~~X_i = X_iZ_i + E_i, Z_i \geq 0,
\end{aligned}
\end{equation}
where
\begin{itemize}
\item $||Z_i||_*$ denotes the low-rank representation revealing the global clustering structure regarding $X_i$.

\item $||Z_i||_1$ aims at extracting the local sparse representation of each data object in $X_i$.

\item $\textmd{Tr}(Z_i^TL_iZ_i)$ characterizes the local manifold structure.

\item $\sum_{i,j \in V}||Z_i - Z_j||_2^2$ characterizes the agreement among the sparse and low-rank representations from all $V$ views.

\item $||E_i||_1$ models the possible Laplacian noise contained by $X_i$, we pose $\ell_1$ on $||E_i||$ for noise robustness.

\item $Z_i \geq 0$ is a non-negative constraint to ensure that each data object is amid its neighbors, through $X_i = X_iZ_i + E_i$, so that the data correlations can be well encoded for the $i^{th}$ view.

\item $\lambda_{1}, \lambda_{2}, \lambda_{3}, \beta$ are all trade-off parameters

\end{itemize}

Eq.\eqref{eq:object} is a typical low-rank optimization problem, and a lot of methods are available to solve it. Among these methods, the Alternating Direction Method is the typical solution, which aims at updating each variable alternatively by minimizing the augmented lagrangian function in a Gauss-Seidel fashion. In this paper, we deploy the method of \textbf{L}inearized \textbf{A}lternating \textbf{D}irection \textbf{M}ethod with \textbf{A}daptive \textbf{P}enalty, dubbed \textbf{LADMAP} \cite{LinNIPS2011}. The underlying philosophy of \textbf{LADMAP} is to linearly represent the smooth component, which enables Lagrange multipliers to be updated within the feasible approximation error.

Observing that solving all the $Z_i, E_i (i \in V)$ pairs follows the same type of optimization strategy, we only present the optimization strategy for the $i^{th}$ view. To resolve this, we first introduce an auxiliary variable $G_i$, then solving the Eq.\eqref{eq:object} with respect to $Z_i, E_i$ and $G_i$ can be written as follows
\begin{equation}\label{eq:variation}
\begin{aligned}
& \min_{Z_i, E_i, G_i} ||Z_i||_* + \lambda_{1}||E_i||_1 + \lambda_2||G_i||_1   \\
& + \lambda_{3}\textmd{Tr}(Z_i^TL_iZ_i) + \frac{\beta}{2}\sum_{j \in V, j \neq i}||Z_i - Z_j||^2_2 \\
& ~~~~~\textmd{s.t.}~~~~~ X_i = X_iZ_i + E_i, G_i = Z_i, G_i \geq 0.
\end{aligned}
\end{equation}
We then present the augmented lagrangian function of Eq.\eqref{eq:variation} below
\begin{equation}\label{eq:argumented}
\begin{aligned}
& \mathcal{L}(Z_i, E_i, G_i, K_1^{i}, K_2^{i}) \\
& = ||Z_i||_* + \lambda_{1}||E_i||_1 + \lambda_{2}||G_i||_1 + \lambda_{3}\textmd{Tr}(Z_i^TL_iZ_i)\\
& + \frac{\beta}{2}\sum_{j \in V, j \neq i}||Z_i - Z_j||^2_2 + \langle K_1^{i}, X_i - X_iZ_i - E_i \rangle + \\
&\langle K_2^{i}, Z_i - G_i \rangle + \frac{\mu}{2}(||X_i - X_iZ_i - E_i||_2^2 + ||Z_i - G_i||_2^2),
\end{aligned}
\end{equation}
where $K_1^{i} \in \mathbb{R}^{d_i \times n}$ and $K_2^{i} \in \mathbb{R}^{n \times n}$ are Lagrange multipliers, $\langle\cdot,\cdot\rangle$ is the inner product and $\mu > 0$ is a penalty parameter. We update each of the above variables alternatively by minimizing Eq.\eqref{eq:argumented} while with other variables fixed. In what follows, we will provide the details of optimizing Eq.\eqref{eq:argumented} with respect to each variable in next section.

\section{Optimization Strategy}
\subsection{Updating $Z_i$}
Minimizing Eq.\eqref{eq:argumented} w.r.t. $Z_i$ is equivalent to minimizing the following
\begin{equation}\label{eq:Zi}
\begin{aligned}
& \mathcal{L}_1 = ||Z_i||_* + \lambda_{3}\textmd{Tr}(Z_i^TL_iZ_i) + \frac{\beta}{2}\sum_{j \in V, j \neq i}||Z_i - Z_j||^2_2 \\
&  + \frac{\mu}{2}||X_i - X_iZ_i - E_i + \frac{1}{\mu}K_1^{i}||_2^2 + \frac{\mu}{2}||Z_i - G_i + \frac{1}{\mu}K_2^{i}||_2^2
\end{aligned}
\end{equation}
It cannot yield a closed form throughout Eq.\eqref{eq:Zi}. Thanks to \textbf{LADMAP}, we can approximately reconstruct the smooth terms of $\mathcal{L}_1$ via a linear manner. The smooth terms of $\mathcal{L}_1$ are summarized below
\begin{equation}\label{eq:smooth}
\begin{aligned}
& \mathcal{Q}_l(Z_i, E_i, G_i, K_1^{i},K_2^{i}) \\
& = \lambda_{3}\textmd{Tr}(Z_i^TL_iZ_i) + \frac{\beta}{2}\sum_{j \in V, j \neq i}||Z_i - Z_j||^2_2 \\
& + \frac{\mu}{2}||X_i - X_iZ_i - E_i + \frac{1}{\mu}K_1^{i}||_2^2 + \frac{\mu}{2}||Z_i - G_i + \frac{1}{\mu}K_2^{i}||_2^2
\end{aligned}
\end{equation}
Based on Eq.\eqref{eq:smooth}, we convert the problem of minimizing Eq.\eqref{eq:Zi} to minimize Eq.\eqref{eq:equivalent} below
\begin{equation}\label{eq:equivalent}
\mathcal{L}_1 = ||Z_i||_* + \langle\frac{\partial \mathcal{Q}_l}{\partial Z_k}, Z_i - Z_i(k)\rangle + \frac{\xi}{2}||Z_i - Z_i(k)||_2^2,
\end{equation}
where $\frac{\partial \mathcal{Q}_l}{\partial Z_i(k)}$ denotes the partial gradient of $\mathcal{Q}_l$ w.r.t. Z at $Z_i(k)$, and $\mathcal{Q}_l$ is approximated by the linear representation $\langle\frac{\partial \mathcal{Q}_l}{\partial Z_k}, Z_i - Z_i(k)\rangle$ w.r.t. $Z_i(k)$ together with a proximal term $\frac{\xi}{2}||Z_i - Z_i(k)||_2^2$. The above replacement is valid provided that $\xi > 2\lambda_3e(L_i) + \mu(1 + ||X_i||_2^2)$, where $e(L_i)$ denotes the largest eigenvalue of $L_i$. Then the following closed form holds for Eq.\eqref{eq:equivalent} for each update.
\begin{equation}\label{eq:Zifinal}
Z_i = \Theta_{\frac{1}{\xi}}(Z_i(k) - \frac{\partial \mathcal{Q}_l}{\partial Z_i(k)} \cdot \frac{1}{\xi}),
\end{equation}
where $\Theta_{\epsilon} (A) = US_{\epsilon}(\Sigma)V^T$ represents the Singular Value Threshold (SVT) operation.  \textmd{$U \Sigma V^T$} is the singular value decomposition of matrix $A$, and $S_{\epsilon}(x) = sign(x)max(||x|| - \epsilon, 0)$ is called the soft threshold operator, $sign(x)$ is 1 if it is positive and 0 otherwise.
\subsubsection{Insights for Iterative Views Agreement.}
We remark that the intuitions for iterative (during each iteration) views-agreement can be captured by expanding $\frac{\partial \mathcal{Q}_l}{\partial Z_i(k)}$ below
\begin{equation}\label{eq:separable}
\begin{aligned}
& \frac{\partial \mathcal{Q}_l}{\partial Z_i(k)} = \lambda_3Z_i(k)(L_i^T + L_i) \\
& + \mu X_i^T(X_iZ_i(k) - X_i + E_i - \frac{1}{\mu}K_1^{i}) \\
& + \mu(Z_i(k) - G_i + \frac{1}{\mu}K_2^{i}) + \beta \underbrace{\sum_{j \in V, j \neq i} ||Z_i(k) - Z_j||}_\text{$\sum_{j \in V, j \neq i}Z_i(k) - \sum_{j \in V, j \neq i}Z_j$}
\end{aligned}
\end{equation}
We expand the last term $\sum_{j \in V, j \neq i} ||Z_i(k) - Z_j|| = \sum_{j \in V, j \neq i}Z_i(k) - \sum_{j \in V, j \neq i}Z_j$, then we re-write Eq.\eqref{eq:separable} below
\begin{equation}\label{eq:sep1}
 \frac{\partial \mathcal{Q}_l}{\partial Z_i(k)} = C_i -\beta \sum_{j \in V, j \neq i} Z_j,
\end{equation}
where
\begin{equation}\label{eq:Ci}
\begin{aligned}
& C_i = \lambda_3Z_i(k)(L_i^T + L_i) + \mu X_i^T(X_iZ_i(k)\\ 
& - X_i + E_i   \frac{1}{\mu}K_1^{i}) + \\
& \mu(Z_i(k) - G_i + \frac{1}{\mu}K_2^{i}) + \beta\sum_{j \in V, j \neq i}Z_i(k) \nonumber
\end{aligned}
\end{equation}
Substituting Eq.\eqref{eq:sep1} into Eq.\eqref{eq:Zifinal} yields Eq.\eqref{eq:mutual}
\begin{equation}\label{eq:mutual}
\begin{aligned}
& Z_i = \Theta_{\frac{1}{\xi}}(Z_i(k) - (C_i -\beta \sum_{j \in V, j \neq i} Z_j) \cdot \frac{1}{\xi})\\
& = \Theta_{\frac{1}{\xi}}(Z_i(k) - \frac{C_i}{\xi} + \beta \underbrace{\sum_{j \in V, j \neq i} Z_j}_\text{Influence from other views} \cdot \frac{1}{\xi}),
\end{aligned}
\end{equation}
where $Z_i$ updating is explicitly influenced from other views \ie $\sum_{j \in V, j \neq i} Z_j$, which reveals that such low-rank representations \eg $Z_i$ updating from each view \eg the $i^{th}$ view are formed by referring to the other views, while served as a constraint to update other views for each iteration so that the complementary information from all views are intuitively leveraged towards a final agreement for clustering.

\subsection{Updating $E_i$}
Minimizing Eq.\eqref{eq:argumented} w.r.t. $E_i$ is equivalent to solving the following optimization problem
\begin{equation}\label{eq:Ei}
\min_{E_i} \lambda_1 ||E_i||_1 + \frac{\mu}{2}||E_i - (X_i - X_iZ_i + \frac{1}{\mu}K_1^{i})||_2^2,
\end{equation}
where the following closed form solution is hold for $E_i$ according to \cite{CaiSIAMJ08}
\begin{equation}\label{eq:EiS}
E_i = S_{\frac{\lambda_1}{\mu}}(X_i - X_iZ_i + \frac{1}{\mu}K_1^{i})
\end{equation}

\subsection{Updating $G_i$}
Minimizing Eq.\eqref{eq:argumented} w.r.t. $G_i$ is equivalent to solving the following optimization problem
\begin{equation}\label{eq:Gi}
\min_{G_i \geq 0} \lambda_2 ||G_i||_1 + \frac{\mu}{2}||G_i - (Z_i + \frac{1}{\mu}K_2^{i})||_2^2,
\end{equation}
where the following closed form solution holds for $G_i$ according to \cite{CaiSIAMJ08}
\begin{equation}\label{eq:GiS}
G_i = \textmd{max}\{S_{\frac{\lambda_2}{\mu}}(Z_i + \frac{1}{\mu}K_2^{i}), 0\}
\end{equation}

\subsection{Updating $K_1^{i}$ and $K_2^{i}$}
We update Lagrange multipliers $K_1^{i}$ via
\begin{equation}\label{eq:K1s}
K_1^{i} = K_1^{i} + \mu(X_i - X_iZ_i - E_i)
\end{equation}
and $K_2^{i}$ via
\begin{equation}\label{eq:K1s}
K_2^{i} = K_2^{i} + \mu(Z_i - G_i)
\end{equation}
We remark that $\mu$ can be tuned using the adaptive updating strategy as suggested by \cite{LinNIPS2011} to yield a faster convergence. The optimization strategy alternatively update each variable while fixing others until the convergence condition is met.

Thanks to \textbf{LADMAP} \cite{LinNIPS2011}, the above optimization process converges to a globally optimal solution enjoyed. Besides, we may employ the Lanczos method to compute the largest singular values and vectors by only performs multiplication of  $Z_i(k) - \frac{\partial \mathcal{Q}_l}{\partial Z_i(k)} \cdot \frac{1}{\xi}$ with vectors, which can be efficiently computed by such successive matrix-vector multiplications.

\subsection{Clustering with $Z_i(i = 1,\ldots,V)$}
Once the converged $Z_i(i = 1,\ldots,V)$ are learned for each of the $V$ views, we normalize all column vectors of $Z_i(i = 1,\ldots,V)$ while set small entries under given threshold $\tau$ to be 0. After that, we can calculate the similarity matrix $W_i (j,k) =\frac{Z_i(j,k) + Z_i(k,j)}{2}$ for the $i^{th}$ view between the $j^{th}$ and $k^{th}$ data objects. The final data similarity matrix for all views are defined as
\begin{equation}\label{eq:simi-all}
W = \frac{\sum_{i}^{V}W_i}{V}
\end{equation}
The spectral clustering is performed on $W$ calculated via Eq.\eqref{eq:simi-all} to yield the final multi-view spectral clustering result.
\section{Experiments}
We evaluate our method on the following data sets:

\begin{itemize}
\item \underline{UCI handwritten Digit set}\footnote{http://archive.ics.uci.edu/ml/datasets/Multiple+Features}: It consists of features of hand-written digits (0-9). The dataset is represented by 6 features and contains 2000 samples with 200 in each category. Analogous to \cite{LinNIPS2011}, we choose 76 Fourier coefficients  (FC) of the character shapes and the 216 profile correlations (PC) as two views.

\item \underline{Animal with Attribute} (AwA)\footnote{http://attributes.kyb.tuebingen.mpg.de}: It consists of 50 kinds of animals described by 6 features (views): Color histogram ( CQ, 2688-dim),  local self-similarity (LSS, 2000-dim),  pyramid HOG (PHOG, 252-dim), SIFT (2000-dim), Color SIFT (RGSIFT, 2000-dim), and SURF (2000-dim). We randomly sample 80 images for each category and get 4000 images in total.

\item \underline{NUS-WIDE-Object (NUS)} \cite{NUS-Wide}: The data set consists of 30000 images from 31 categories. We construct 5 views using 5 features as provided by the website \footnote{lms.comp.nus.edu.sg/research/NUS-WIDE.html}: 65-dimensional color histogram (CH), 226-dimensional color moments (CM), 145-dimensional color correlation (CORR), 74-dimensional edge estimation (EDH), and 129-dimensional wavelet texture (WT).
\end{itemize}

These data sets are summarized in Table \ref{table:dataset}.

\begin{table}[t]
\caption{Summary of the multi-view data sets used in our experiments.}
\begin{tabular}{cccc}
\hline
Features&UCI & AwA& NUS\\
\hline
1 & FC (76) & CQ (2688) & CH(65)\\
2 & PC (216) & LSS (2000) & CM(226)\\
3 & - & PHOG (252)& CORR(145)\\
4 & - &SIFT(2000) & EDH(74)\\
5 & -& RGSIFT(2000) & WT(129)\\
6 & - & SURF(2000) & -\\
\hline
\# of data & 2000 & 4000 & 26315\\
\# of classes & 10 & 50 & 31\\
\hline
\end{tabular}
\label{table:dataset}
\end{table}

\subsection{Baselines}
We compare our approach with the following state-of-the-art baselines:

\begin{itemize}
\item \textbf{MFMSC}: Using the concatenation of multiple features to perform spectral clustering.

\item Multi-view affinity aggregation for multi-view spectral clustering (\textbf{MAASC}) \cite{CVPR12}.

\item Canonical Correlation Analysis (CCA) based multi-view spectral clustering (\textbf{CCAMSC}) \cite{ICML09}: Projecting multi-view data into a common subspace, then perform spectral clustering.

\item Co-regularized multi-view spectral clustering (\textbf{CoMVSC}) \cite{NIPS11}: It regularizes the eigenvectors of view-dependent graph laplacians and achieve consensus clusters across views.

\item \textbf{Co-training} \cite{icml11}: Alternately modify one view's eigenspace of graph laplacian by referring to the other views' graph laplacian and their corresponding eigenvectors, upon which, the spectral clustering is conducted. Such process is performed until convergence.

\item Robust Low-Rank Representation method (\textbf{RLRR}) \cite{RMVSC}, as formulated in Eq.\eqref{eq:intro}.
\end{itemize}

\subsection{Experimental Settings and Parameters Study}

For fair comparison, we implement these competitors by following their experimental  setting and the parameter tuning steps in their papers. The Gaussian kernel is used throughout experiments on all data sets and $\sigma$  in Eq.\eqref{eq:gaussian} is learned by self-tuning method \cite{nips04}, and
$s=20$ to construct $s$-nearest neighbors for each data object to calculate Eq.\eqref{eq:gaussian}.
To measure the clustering results, we use two standard metrics: clustering accuracy (\textsf{ACC}) (Ratio for the number of data objects having same clustering label and ground truth label against total data objects), and normalized mutual information (\textsf{NMI}). Pleaser refer to \cite{Chen-TPAMI11} for details of these two clustering metrics. All experiments are repeated 10 times, and we report their averaged mean value.\\
\textbf{\underline{Feature noise modeling for robustness}}: Following \cite{CV155}, for each view-specific feature representation, 20\% feature elements are corrupted with uniform distribution over the range [5,-5], which is consistent to the practical setting while matching with \textbf{RLRR} and our method.

We set $\lambda_1=2$, $\lambda_2=0.08$ in Eq.\eqref{eq:object}. \emph{To validate the effectiveness of Multi-graph regularization and iterative views agreement, we test the value of \textsf{ACC} and \textsf{NMI} over a range of $\lambda_3$ and $\beta$ in Eq.\eqref{eq:object} in the next subsection.}

\subsection{Validating Multi-graph regularization and Iterative views agreement}
\begin{figure}[t]
\begin{tabular}{c}
\includegraphics[width=8cm, trim=250 10 30 28, clip]{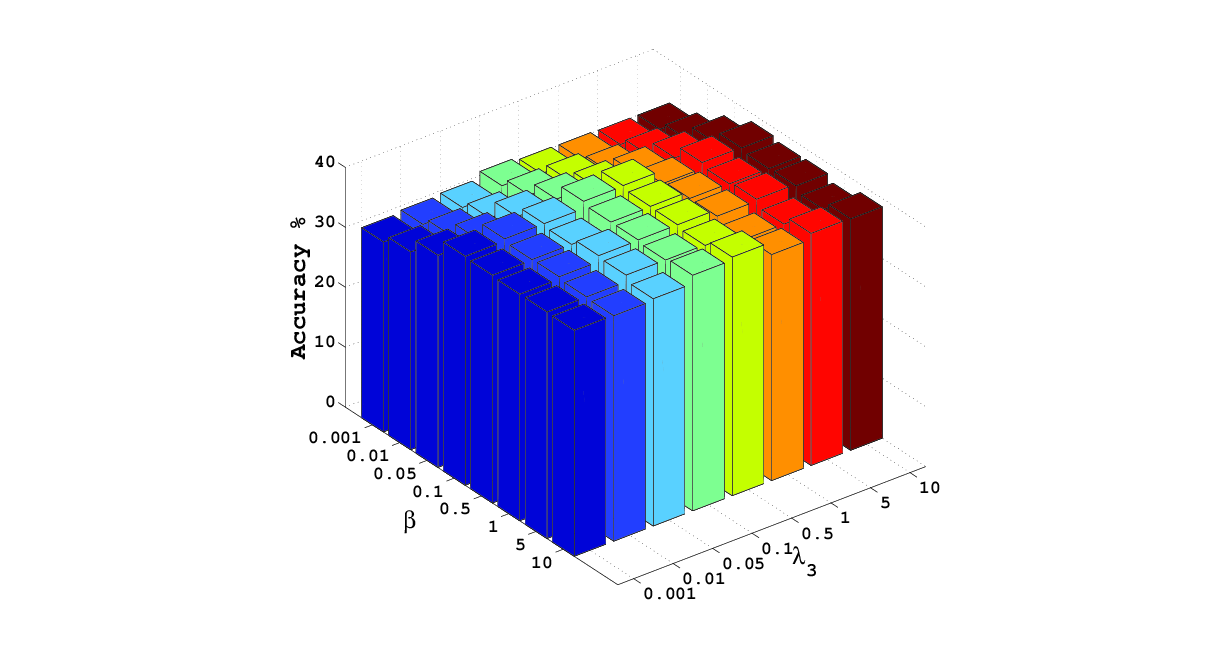}\\
(a)\\
\includegraphics[width=8cm, trim=200 10 30 28, clip]{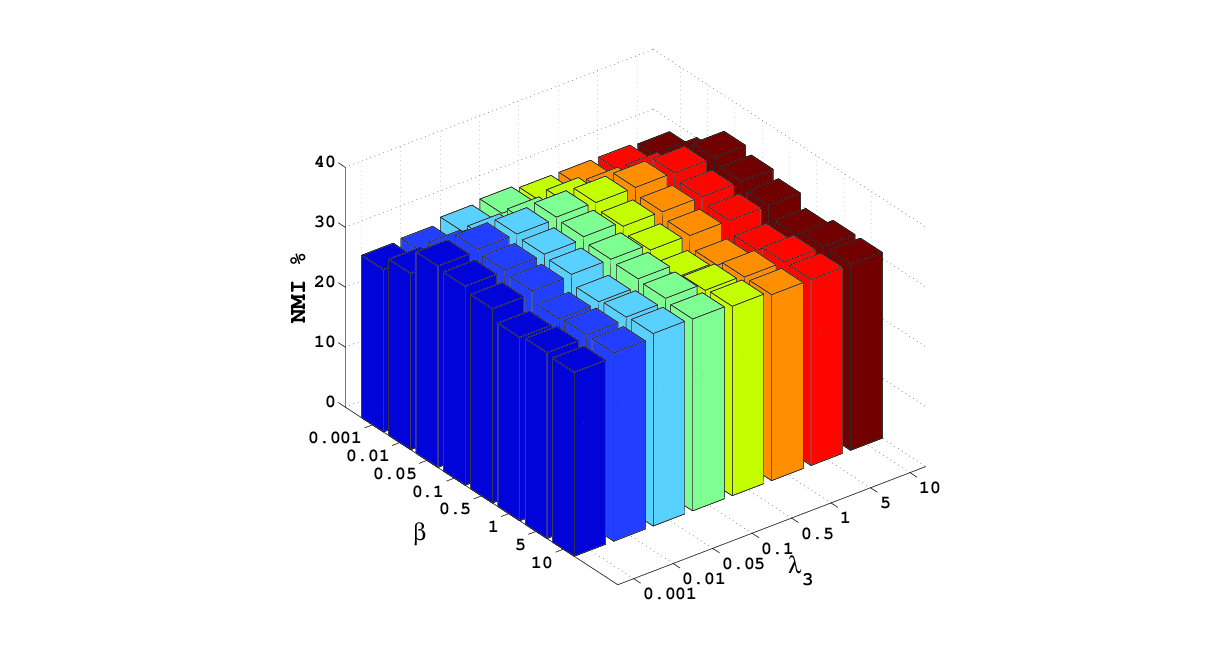}\\
(b)
\end{tabular}
\caption{Parameters study of $\lambda_3$ and $\beta$ for our multi-graph regularization and iterative views agreement scheme on AwA dataset. (a) \textsf{ACC} against parameters $\lambda_3$ and $\beta$. (b) \textsf{NMI} against parameters $\lambda_3$ and $\beta$.}\label{fig:parameter}
\end{figure}

We test $\lambda_3$ for multi-graph regularization term and $\beta$ for iterative views agreement within the interval [0.001,10]  over the AwA data set and adopt such setting for other data sets. Specifically, we test each value of one parameter while fixing the value of the other parameter, the results are then illustrated in Fig. \ref{fig:parameter}.

From both Fig.\ref{fig:parameter} (a) and (b), the following observations can be identified:

\begin{itemize}
\item when fixing the value of $\lambda_3$,  increasing the value of $\beta$ can basically improves the \textsf{ACC} and \textsf{NMI} value of our method. The similar observation can be identified vice versa; that is, fixing the value of $\beta$, meanwhile increasing the $\lambda_3$ can always lead to the clustering improvement in terms of both \textsf{ACC} and \textsf{NMI}.
\item Both the above clustering measures \textsf{ACC} and \textsf{NMI} will unsurprisingly increase when both $\lambda_3$ and $\beta$ increases until reach the optimal pair-combinations, then slightly decrease.
\end{itemize}
Upon the above observations,  we choose a balance pair values: $\lambda_3=0.5$ and $\beta=0.1$ for our method.
\subsection{Experimental Results and Analysis}

\begin{table}[t]
\caption{Clustering results in terms of \textsf{ACC} on three benchmark data sets.}
\begin{tabular}{cccc}
\hline\hline
\textsf{ACC} (\%) & UCI digits & AwA & NUS \\
\hline
\textbf{MFMSC} & 43.81 & 17.13& 22.81\\
\textbf{MAASC} & 51.74 & 19.44& 25.13\\
\textbf{CCAMSC} & 73.24 & 24.04& 27.56\\
\textbf{CoMVSC} & 80.27 & 29.93 & 33.63\\
\textbf{Co-training} & 79.22 & 29.06 & 34.25\\
\textbf{RLRR} & 83.67 & 31.49 & 35.27\\
\textbf{Ours} & \textbf{86.39} & \textbf{37.22} & \textbf{41.02}\\
\hline
\end{tabular}
\label{table:acc}
\end{table}

\begin{table}[t]
\caption{Clustering results in terms of \textsf{NMI} on three benchmark data sets.}
\begin{tabular}{cccc}
\hline\hline
\textsf{NMI} (\%) & UCI digits & AwA & NUS \\
\hline
\textbf{MFMSC} & 41.57 & 11.48 & 12.21\\
\textbf{MAASC} & 47.85 & 12.93& 11.86\\
\textbf{CCAMSC} & 56.51 & 15.62& 14.56\\
\textbf{CoMVSC} & 63.82 & 17.30 & 7.07\\
\textbf{Co-training} & 62.07 & 18.05 & 8.10\\
\textbf{RLRR} & 81.20 & 25.57 & 18.29\\
\textbf{Ours} & \textbf{85.45} &\textbf{31.74} & \textbf{20.61}\\
\hline
\end{tabular}
\label{table:nmi}
\end{table}

We report the compared clustering results in terms of \textsf{ACC} and \textsf{NMI} in
Table \ref{table:acc} and Table \ref{table:nmi},  upon which, the following observations can be drawn:

\begin{itemize}
\item Nearly most of the clustering performance in terms of both \textsf{ACC} and \textsf{NMI} are better than \textbf{RLRR}, which further demonstrates the effectiveness of our multi-graph regularization and iterative views agreement scheme combining with LRR scheme for multi-view spectral clustering.

\item First, comparing with view-fusion methods like \textbf{MFMSC} and \textbf{MAASC}, our method improves the clustering performance by a notable margin on all data sets. Specifically, it highly improves the clustering performance in terms of \textsf{ACC} from 43.81\% (\textbf{MFMSC}), 51.74\% (\textbf{MAASC}) to 86.39\% on UCI digits data set.  Such notable improvement can be also observed on AwA and NUS data sets.

\item Second, \textbf{CCAMSC} that learns a common low-dimensional subspace among multi-view data is less effective in clustering due to its incapability of encoding local graph structures from heterogeneous views within only a common subspace. In contrast, our method can well address such problem with a novel iterative views-agreement scheme, which is notably evidenced in terms of both \textsf{ACC} and \textsf{NMI}.

\item Comparing with co-regularized paradigms (\textbf{CoMVSC}, and \textbf{Co-training}), our method works more effectively in the presence of noise corruptions. For example, in NUS data set, it improves the clustering accuracy from 33.63\%(\textbf{CoMVSC}), 34.25\% (\textbf{Co-training})  to 41.02\%.  Although \textbf{RLRR} is also effective to deal with practical noise-corrupted multi-view data. However, as aforementioned, learning only one common low-rank correlation representation shared by all views is failed to flexibly capture all the local nonlinear manifold structures from all views, which is crucial to spectral clustering, while our technique can deliver a better performance.
\end{itemize}

\section{Conclusions}
In this paper, we propose an iterative structured low-rank optimization method to multi-view spectral clustering. Unlike existing methods, Our method can well encode the local data manifold structure from each view-dependent feature space, and achieve the multi-view agreement via an iterative fashion, while better preserve the flexible nonlinear manifold structure from all views. The superiorities are validated by extensive experiments over real-world multi-view data sets.

One future direction is to adapt the proposed iterative fashion technique to cross-view based research \cite{YangMM13,YangSIGIR} by dealing with multiple data source yet corresponding to the same latent semantics. We aim to develop the novel iterative technique to learn the projections for multiple data sources into the common latent space to well characterize the shared latent semantics.

\bibliographystyle{named}
\bibliography{ijcai16}

\end{document}